\documentclass{article}

\usepackage{arxiv}

\usepackage[utf8]{inputenc} 
\usepackage[T1]{fontenc}    
\usepackage{hyperref}       
\usepackage{url}            
\usepackage{booktabs}       
\usepackage{amsfonts}       
\usepackage{nicefrac}       
\usepackage{microtype}      
\usepackage{lipsum}		
\usepackage{graphicx}
\usepackage{natbib}
\usepackage{doi}

\usepackage{graphicx}
\usepackage{amsmath}
\usepackage{amssymb}
\usepackage{booktabs}
\usepackage{algorithm}
\usepackage{algpseudocode}
\usepackage{xcolor}

\algnewcommand\algorithmicforeach{\textbf{for each}}
\algdef{S}[FOR]{ForEach}[1]{\algorithmicforeach\ #1\ \algorithmicdo}
\algdef{SE}[SUBALG]{Indent}{EndIndent}{}{\algorithmicend\ }%
\algtext*{Indent}
\algtext*{EndIndent}

\title{Faithful Counterfactual Visual Explanations (FCVE)}

\date{} 					

\author{ Bismillah Khan \\
Department of Computer Science\\
COMSATS University Islamabad\\
Islamabad, Pakistan \\
\texttt{bkhan1903@gmail.com} \\
\And
Syed Ali Tariq \\
	Department of Computer Science\\
	COMSATS University Islamabad\\
	Islamabad, Pakistan \\
	\texttt{s.alitariq1@gmail.com} \\
	\And
	Tehseen Zia \\
	Department of Computer Science\\
	COMSATS University Islamabad\\
	Islamabad, Pakistan \\
	\texttt{tehseen.zia@comsats.edu.pk} \\
	\AND
	Muhammad Ahsan \\
	The City School \\
	Ravi Campus \\
	Pakistan \\
	\texttt{ahsan.warraich200@gmail.com} \\
	\AND
	David Windridge \\
	Middlesex University London \\
	UK \\
	\texttt{d.windridge@mdx.ac.uk} \\
}

\begin{document}
\maketitle

\begin{abstract}
	Deep learning models in computer vision have made remarkable progress, but their lack of transparency and interpretability remains a challenge. The development of explainable AI can enhance the understanding and performance of these models. However, existing techniques often struggle to provide convincing explanations that non-experts easily understand, and they cannot accurately identify models' intrinsic decision-making processes. To address these challenges, we propose to develop a counterfactual explanation (CE) model that balances plausibility and faithfulness. This model generates easy-to-understand visual explanations by making minimum changes necessary in images without altering the pixel data. Instead, the proposed method identifies internal concepts and filters learned by models and leverages them to produce plausible counterfactual explanations. The provided explanations reflect the internal decision-making process of the model, thus ensuring faithfulness to the model.
\end{abstract}


\section{Introduction}

Deep convolution neural networks (DCNNs) are at the leading edge of technology in many advanced areas of computer vision applications such as healthcare \cite{uddin2018activity}, criminal justice \cite{schiliro2021novel}, banking finance decisions \cite{abakarim2018towards}, transportation \cite{alfarraj2020internet}, agriculture \cite{zhang2018identification}, fraud detection \cite{chouiekh2018convnets} and scene segmentation \cite{seijdel2020depth}, etc. The extensive use of deep convolutional neural networks over a conventional neural network is due to the fact that they are computationally competitive, automatically learn a hierarchy of representations from the input data \cite{li2021survey}, and are agile compared to neural networks \cite{nandhini2021deep}. 

However, DCNNs are opaque in nature as their innards are not properly understood and visible, making them a black-box \cite{arrieta2020explainable}. The DCNN models need to be transparent for safety-critical applications such as healthcare and criminal justice that involve dealing with human life \cite{tjoa2020survey, holzinger2017we} and driverless vehicles \cite{zablocki2021explainability}, etc., in which the effect of inaccurate or undesired decisions have significant consequences \cite{samek2017explainable, goebel2018explainable, rudin2019stop}. Several studies show that DCNNs often regard dataset bias \cite{zhang2018examining} and rely on undesired or inappropriate features to take decisions. The DCNNs also produce incorrect results when subtle changes are made to the input \cite{akhtar2018threat}. Adversarial attacks cause risk to several security-critical applications, for instance, in driver-less vehicles where slight obstructions on traffic signs can result in undesired conclusions \cite{eykholt2018robust} or in surveillance systems where malevolent individuals may cause harm \cite{thys2019fooling}. Therefore, DCNNs are unreliable and need explainable AI (XAI) approaches to determine their deficiencies and train trustworthy, robust, and transparent models \cite{rudin2019stop, ghorbani2020neuron, fong2017interpretable}.

Different types of (XAI) methods exist in the literature and can be categorized into two dominant groups: ante-hoc \cite{du2019techniques} and post-hoc \cite{vale2022explainable}. Although the ante-hoc models have intrinsically explainable model structure, the explainability comes at the cost of lower performance. The post-hoc models tend to explain other pre-build black-box models; hence they do not compromise performance at the expense of explainability. Among many post hoc techniques, counterfactual and contrastive explanations have emerged as powerful visual explanation types.

Contrastive explanations usually identify the actual features of the input data that play an important role in model decision-making for the inferred class \cite{dhurandhar2018explanations}. Such explanations are meaningful as they imitate the process of human thoughts and are easily understood. Counterfactual explanations describe what features need to be modified and to what extent to flip the decision of the model (i.e., to reverse an undesired outcome). Counterfactual explanations offer recourse by trying to find the minimum change in the input data to obtain a positive result \cite{wachter2017counterfactual, karimi2020model,poyiadzi2020face,van2021interpretable}. On the basis of such explanations, we come across the reasons behind the model predictions; hence we can either accept or reject the given prediction accordingly. Several contrastive and counterfactual explanation methods have been proposed recently \cite{pmlr-v97-goyal19a, hendricks2018grounding,dhurandhar2018explanations,selvaraju2017grad, luss2019generating}, in which certain input data pixels are perturbed to alter the model's prediction.

However, a critical shortcoming of these existing approaches is that they are not faithful (or aligned) to the model and do not make the model transparent (i.e. glass-box, rather than black-box) in terms of its reasoning process. Further, these methods aim to find the optimal combination of pixels for perturbing or in-filling, and they are computationally expensive \cite{poyiadzi2020face}. Addressing the issue, a recent study \cite{akula2020cocox} deals with super-pixels rather than pixels to find crucial decisive concepts that, when deleted from or added to the query image, affect the model's decision. Despite generating useful explanations, this method is not faithful and glass-box transparent as it generates explanations by operating on pixel data. Another line of research aims to identify whether a particular concept has some significance to a given model \cite{pmlr-v80-kim18d}. This approach, however, neither investigates the internal reasoning of DCNN models nor provides counterfactual and contrastive explanations. 

This research is based on a recent study \cite{TARIQ2022109901}, which deals with identifying counterfactuals and contrastive filters of DCNN models rather than pixels of an image. Despite generating counterfactual and contrastive filters, the approach does not provide visual explanations of the generated counterfactual and contrastive ones. Hence, this study aims to suggest a post-hoc explainability method that visually explains the predictive identification of counterfactuals and contrastive filters in a DCNN model.

In this regard, the proposed solution is a Faithful Counterfactual Visual Explanation (FCVE) model.

\section{Related Work}
\label{sec:two}

The authors in \cite{pmlr-v97-goyal19a} propose a method for generating counterfactual visual explanations to provide insights into the decision-making process of deep learning models. The authors employ GANs to generate alternative images that would have led to different model predictions. They use a conditional GAN framework where the generator is conditioned on the input image and a desired output class. While the generated counterfactuals are visually plausible, the evaluation of faithfulness, i.e., the degree to which the generated explanations accurately reflect the model's internal reasoning, is not extensively discussed. In the paper \cite{khorram2022cycle} authors introduce an intriguing approach to generate counterfactual explanations using latent space transformations. The authors propose a method that leverages the power of generative models, specifically CycleGAN, to produce counterfactual instances by mapping an original instance to a counterfactual representation in the latent space and then back to the input space. This cycle-consistency constraint ensures that the generated counterfactuals retain important features of the original instance while introducing meaningful modifications. The paper not only focuses on generating counterfactuals but also discusses their utility in enhancing interpretability and fairness in machine learning models. This broader perspective strengthens the paper's significance and relevance in addressing the need for explainable AI systems. While the paper presents an innovative and promising approach, it also has some limitations worth considering. One limitation is the reliance on CycleGAN, which may not capture all the complexities of the original input space. Exploring alternative generative models or incorporating additional constraints could further improve the fidelity and relevance of the generated counterfactual explanations.

The authors in \cite{chang2018explaining} present a method that generates counterfactual explanations for image classifiers. The approach utilizes GANs to generate alternative images by perturbing the input image in a semantically meaningful way. By incorporating contrastive loss and regularization terms, the authors aim to ensure the plausibility and faithfulness of the generated counterfactuals. However, the evaluation of faithfulness is not explicitly addressed, and more rigorous analysis is necessary to determine the extent to which the explanations align with the internal reasoning of the classifier. In \cite{kenny2021generating}, the main focus is on generating plausible counterfactual and semi-factual explanations for deep learning models. The authors propose a method that combines an encoder-decoder architecture with variational autoencoders (VAEs) to generate counterfactual explanations. The generated explanations are evaluated based on their plausibility and faithfulness. The authors provide qualitative analyses and comparisons to demonstrate the faithfulness of their approach, but a more comprehensive quantitative evaluation would further strengthen their claims. Cocox, a framework for generating conceptual and counterfactual explanations, is introduced in \cite{akula2020cocox}. The authors propose a two-step process: first, they learn concept prototypes using GANs, and then generate counterfactual explanations by manipulating latent variables within the GAN framework. While the paper primarily focuses on conceptual explanations, the faithfulness of the generated counterfactual explanations is not explicitly discussed or evaluated. The article \cite{vandenhende2022making} addresses the challenge of generating semantically consistent visual counterfactual explanations. The authors aim to generate plausible counterfactual images that maintain semantic coherence, ensuring that changes to the image do not introduce unrealistic or incoherent elements. The study presents a novel framework that leverages GANs to generate semantically consistent visual counterfactuals. The authors propose a two-step approach consisting of modification and regularization phases. In the modification phase, they use a conditional GAN to generate counterfactual images by introducing changes to the original image. The GAN is trained to preserve the image semantics while incorporating user-specified changes. The regularization phase involves a semantic consistency loss term that encourages the generated images to maintain semantic coherence throughout the modification process. The authors evaluate their framework using qualitative and quantitative assessments. They compare their method with existing approaches and demonstrate that it produces visually realistic and semantically consistent counterfactual images. They perform user studies to measure the generated counterfactuals' perceived plausibility and semantic coherence, obtaining favorable results. The paper addresses an important aspect of counterfactual explanation generation, emphasizing the need for explanations that align with human perception and understanding. The authors in \cite{lang2021explaining} present an approach that revolutionizes the field of interpretable machine learning. By combining Generative Adversarial Networks (GANs) and StyleSpace analysis, they introduce a method that generates visually captivating explanations for classifier decisions. The authors demonstrate the efficacy of their framework by manipulating the latent space of a GAN to create images that clarify the underlying rationale behind a classifier's output. The disentangled properties of StyleGAN enable the generation of interpretable visual attributes, showcasing the ability of the proposed method to capture essential features driving classifier decisions.

The novelty of this paper lies in its fusion of GANs and StyleSpace analysis to produce explanations that surpass conventional textual justifications. By exploiting the unique characteristics of StyleGAN, the authors unlock the potential to manipulate specific visual attributes within the generator's latent space. This approach allows for the creation of visually intuitive explanations that go beyond traditional methods, ensuring that the generated images are both interpretable and relevant. While the paper's reliance on labeled data and its focus on image-based explanations present limitations.
\textcolor{black}{When the classifier exhibits biases or errors, the StylEx may inadvertently captures and amplify these inaccuracies, due to its dependence on the quality of the underlying classifier. Additionally, the performance of StylEx could be impacted while dealing with complex datasets, where attributing changes in classifier decisions to specific visual attributes may be challenging. Moreover, the effectiveness of this method in handling multi-attribute counterfactual explanations decreases.\\
StylEx's limitations could be overcome through the consideration of multiple enhancements. Firstly, The underlying classifier training process can be made more robust and fair attribute extraction by incorporating techniques for bias detection and mitigation. The classifier model could be less susceptible to inaccuracies if it is regularly audited and updated, particularly when there is biased data. Additionally, researchers could concentrate on improving StylEx to better handle counterfactual explanations that involve multiple attributes.
}The authors in \cite{alipour2022explaining} introduce a compelling method for generating counterfactual explanations. By leveraging generative models and enforcing cycle-consistency, the authors provide a valuable contribution to the field of interpretable machine learning. The paper's comprehensive evaluation, along with its focus on interpretability and fairness, highlights its potential impact in improving transparency and trust in AI systems.
\textcolor{black}{However, there are some limitations to consider. When managing very complex generative model latent spaces may be a challenge for the method, to find clear paths for attribute changes. Additionally, accurately training the shift predictor, an important part of the process, can be tricky. These limitations could impact the overall effectiveness.
To address limitations, enhance the method's attribute disentanglement by incorporating advanced techniques such as disentangled representation learning. Validate and generalize the proposed approach across diverse datasets and image classification tasks to ensure broader applicability. Conduct thorough experiments to assess the effectiveness of the pipeline in detecting and mitigating bias in image classification systems under various real-world scenarios.} An innovative approach is presented in \cite{augustin2022diffusion} to generate visual counterfactual explanations using diffusion models. The authors propose a method that leverages the power of diffusion models to transform an input image into a counterfactual representation by iteratively updating the pixel values. By incorporating a contrastive loss function, the generated counterfactual explanations highlight the minimal changes required to alter the classification decision of a deep neural network. The paper provides a thorough evaluation of the proposed method, demonstrating its effectiveness in generating interpretable visual explanations and its potential impact on enhancing transparency and interpretability in deep learning systems. 
\textcolor{black}{Although DVCEs can provide promising insights into image classifier decisions, there are certain limitations that must be taken into consideration. Even with adaptive parameterization, it's still a challenge to produce semantically meaningful changes. DVCEs' effectiveness can be demonstrated by choosing the right hyperparameters, and the optimal selection may vary across different datasets and classifiers. Furthermore, it might require careful parameter tuning. Further investigation is necessary to determine if DVCEs can be applied to a wider range of classifiers and datasets, and the method's sensitivity to variations of input and model architectures should be thoroughly investigated.
To reduce the computational cost of multiple iterations, enhancing computational efficiency can be achieved through optimization and parallelization. Examining alternative denoising methods for insecure models and enhancing approximation techniques to strengthen the theoretical foundations of DVCEs.}
The authors in \cite{balasubramanian2020latent} present an effective baseline method for generating reverse counterfactual explanations. The simplicity of the proposed method, combined with its competitive performance, makes it a valuable addition to the field of interpretable machine learning. Addressing potential biases and exploring the generalizability of the approach would be valuable directions for future research. This paper provides a solid foundation for generating reverse counterfactual explanations and opens avenues for further advancements in this area. 
\textcolor{black}{The method discussed in the paper, called Latent-CF, can be utilized effectively for particular types of data, such as images and loan details. The method might not work as smoothly if the datasets have many different features. In simpler terms, it's like a tool that works well for specific datasets, but we are not entirely sure how it handles different tasks.
Future research should utilize diverse datasets to address limitations, consisting of diverse table and high-dimensional datasets, and conduct a comprehensive investigation of alternative optimization strategies within the latent space, such as genetic algorithms or Bayesian-driven approaches, to enhance the generalizability and robustness of the proposed Latent-CF method.
}
The paper \cite{jeanneret2022diffusion} presents a comprehensive exploration of the application of diffusion models for generating counterfactual explanations. The authors propose an approach that utilizes the power of diffusion models to generate plausible and interpretable counterfactual instances by iteratively updating the input data. The paper provides a thorough analysis of the benefits and limitations of diffusion models in the context of counterfactual explanation generation, highlighting their ability to capture complex data distributions and generate meaningful modifications. The extensive evaluation on various datasets and comparison with existing methods demonstrate the effectiveness and superiority of diffusion models for generating high-quality counterfactual explanations.
\textcolor{black}{Even though DiME has been successful, it is necessary to admit some limitations. The method uses diffusion models that may require substantial processing resources makes the computation costly during inference time, is a significant drawback. The model may be insufficient for applications that need instant interpretation due to the challenges associated with real-time explanations. These limitations need for further research to address computational efficiency and applicability in time-sensitive scenarios.
In order to minimize the limitations of DiME, exploring techniques like model parallelism can help improve computational efficiency or reducing significant inference time through algorithmic optimizations. Simultaneously, by investigating transfer learning, it is possible to reduce the dependence on training data or cross-domain adaptation approaches, enabling efficient generation without requiring extensive data access. The goal of these precision enhancements is to improve DiME's efficiency, scalability, and applicability in real-time situations and environments that value privacy.
}

\section{Proposed methodology}
\label{sec:three}

The proposed study aims to develop a post-hoc visual explainability method that provides plausible and faithful counterfactual visual explanations (FCVE) that are easy to understand and offer reasoning behind model decisions, reflecting the internal working process of the model. To accomplish this, we build upon a previously developed counterfactual explanation (CFE) model in \cite{TARIQ2022109901} that identifies counterfactual filters to explain model decisions. It does this by predicting a set of minimum correct (MC) and minimum incorrect (MI) filters. The MC filters are necessary to maintain the prediction of the image to the original inferred class by the classifier. Mathematically, the MC filters can be denoted as 
\begin{equation}
    F_{MC_i} \in [0,1]^{1\times n},
\end{equation}
where $n$ is the number of filters in the top convolution layer of the classifier model. Values of `1' and `0' indicate whether the corresponding filter is to be active or disabled, respectively, to maintain the prediction to the inferred class.

In contrast, the MI filters are needed to alter the classifier’s decision to a chosen target class. Mathematically, the MI filters can be denoted as
\begin{equation}
    F_{MI_i} \in [\mathbb{R}^+]^{1\times n}.
\end{equation}
Non-zero indexes in $F_{MI_i}$ correspond to the MI filters, and the values at these indexes indicate the magnitude by which the original filter activations are altered to modify the classifier's decision.

The CFE model operates on the last convolution layer of the classifier because these filters have the most impact and represent more abstract, high-level features, concepts, and even whole objects \cite{bau2020understanding, zhou2014object, bau2017network}. In paper \cite{TARIQ2022109901}, it is demonstrated that by enabling, disabling, or modifying these high-level filters in certain ways, it is possible to change the decisions of a pre-trained classifier to either the original inferred class or a chosen alternative class. Importantly, the CFE model probes the internal structure of a deep learning model without altering the input, allowing users to provide faithful explanations aligned with the model’s internal decision-making process. Thus, in the study, we rely on these filters as changes to them may produce plausible visual explanations.

\begin{figure*}[hbt!]
\begin{center}
    \includegraphics[width=1.0\linewidth]{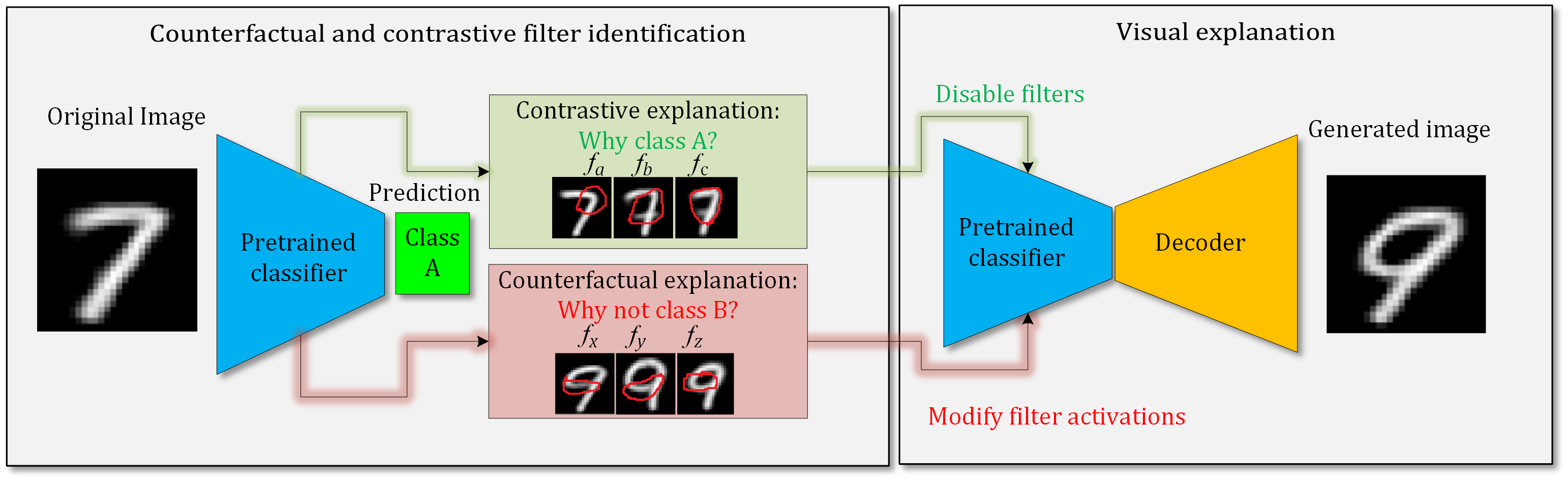}
\end{center}
   \caption{Block diagram of the proposed visual counterfactual explanation model. \textcolor{black}{The proposed method consists of two steps: first is the identification of contrastive and counterfactual filters to explain classifier's decisions, followed by the visualization of these filters by generating images with the modified activations. The decoder is initially trained with all filters intact to recreate the input, so that when the encoder's output is altered using the identified filters, their effect is visualized in the recreated image. }}
\label{fig:1}
\end{figure*}
\subsection{\textbf{\textcolor{black}{Classifier}}} 
To provide visual counterfactual explanations, we propose a joint CFE and classifier-decoder model. In this model, the pre-trained classifier is the model being analyzed, and the decoder model generates visualizations of the classifier's decisions by modifying the filter activations obtained from the CFE model's counterfactual and contrastive explanations. Fig. \ref{fig:1} presents a block diagram illustrating the different phases involved in our proposed model.  

In the counterfactual and contrastive filter identification phase, we extract the MI and MC filters to provide contrastive and counterfactual explanations of the classifier's decisions. These filters capture the necessary changes to alter the classifier's decision to a target class or maintain the original inferred class. In the visual explanation phase, we use a decoder that takes as input the encoded feature vector generated at the last convolution layer of the classifier and it tries to recreate the input image that was given to the classifier. 

The idea behind the decoder model is that it is initially trained to translate the encoded classification features into the respective input while all filters are intact. Once the decoder is trained, we modify the filters using the counterfactual and contrastive explanations produced by the CFE model. This allows us to observe the filter-level changes on the regenerated images, thus generating visual explanations that reflect the alterations made by the counterfactual and contrastive filters.

\textcolor{black}{In the case of Figure \ref{fig:1}, to generate explanation for why a classifier classified an input to class A (i.e., 7 in this case) and not to class B, we can select any target counterfactual class. In this case, we selected class 9 as the counterfactual class due to similarity between them. The proposed method identified counterfactual filters that if they were active in the classification of the input 7, then the model would likely classify the input to the target class. In the case of 7, the proposed method identified key filters responsible for turning the decision from 7 to 9. And when the decoder was presented with 7 as input with the modified filters responsible for classifying 7 as 9, the decoder regenerated the 7 as a 9, demonstrating features that needed to be present in the input image to be classified into class 9.}

To train the proposed visual counterfactual explanation model, we follow a two-phase approach. In the first phase, we train the decoder model with mean absolute error (MAE) loss while the classifier weights are frozen to reproduce the input image given to the classifier. The loss function is defined as follows:
\begin{equation}
\min_{D}\frac{1}{n}\sum_{i=1}^{n}|x_{i}-D(C_{conv}(x_{i}))|_{1},
\label{eq:decoder}
\end{equation}
where $x_{i}$ represents the $i^{th}$ input image, $C_{conv}(x_{i})$ denotes the encoded feature vector produced by the last convolutional layer of the classifier model $C$ for the $i^{th}$ input image, and $D(C_{conv}(x_{i}))$ represents the reconstructed image generated by the decoder model $D$ using the encoded feature vector $C_{conv}(x_{i})$. The mean absolute error loss is calculated as the average of the absolute differences between the input image $x_{i}$ and its corresponding reconstructed image $D(C_{conv}(x_{i}))$ for all $n$ input images. The training process aims to minimize this MAE, thereby ensuring the decoder effectively reproduces the input images.

In the second phase, we utilize pre-trained CFE model to generate MC and MI filters for the given classifier. 
The CFE model can be represented as a function that takes an input image $x$ and produces a set of $F_{MI_i}$ and $F_{MC_i}$ filters as output. This can be expressed as:
\begin{equation}
\label{eq:2}
F_{MI_i}, F_{MC_i} = \text{CFE}(x,C,\hat{c}), 
\end{equation}
where $x$ denotes the input image, $C$ represents the pre-trained classifier, $\hat{c}$ is the target class, and $\text{CFE}$ is the counterfactual explanation model. The $\text{CFE}$ model is responsible for generating the sets of MC and MI filters required to maintain the original classification decision and change it to the target class, respectively, using the following equations 
\begin{equation}\label{eq:new4}
F_{MC_i} = \text{ReLU}_t(\text{Sigmoid}(d^n(g_i))),
\end{equation}
\begin{equation}\label{eq:new6}
F_{MI_i} = \text{ReLU}(d^n(g_i)),
\end{equation}
where $g_i$ denotes the feature maps obtained after the global average pooling layer of classifier $C$, $d^n$ represents a dense layer with $n$ units, $\text{Sigmoid}$ denotes the sigmoid activation function, and $\text{ReLU}_t$ is a thresholded-ReLU layer with a threshold value of $t = 0.5$. The $\text{ReLU}t$ layer produces the approximately binarized MC filter map $F_{MC_i}$ by setting all values below the threshold to zero and leaving the other values unchanged, and $\text{ReLU}$ denotes the rectified linear unit activation function.

These MC and MI filters are utilized to modify the filter activations of the classifier to observe their impact on the reconstructed images. The process can be described by the following equation:\textcolor{black}{
\begin{equation}
C\big(x, F_{MC_i}, F_{MI_i}\big).
\end{equation}
In this equation, the input image $x$, along with the MC and MI filters, are provided as input to the pre-trained classifier $C$. The classifier then generates an altered feature vector by incorporating the effects of these filters.
\subsection{\textbf{Decoder}}
We designed an asymmetric encoder-decoder architecture to synthesize counterfactuals visually. The decoder is asynchronous as the encoder and decoder have variable depths (the number of deconvolution and up-sampling layers of the decoder are not equal to convolution and max pooling layers of the encoder model). The decoder has lower depth than the encoder consequently, the decoder can be trained efficiently. We train the decoder model once the encoder model is trained for the prediction of MC and MI filters. The decoder model reconstructs the latent representation provided by the CFE model. The CFE model working as encoder uses pretrained VGG16 which down sizes the input image x to the last layer as a feature vector in the latent space. The CFE model learns the extent of changes to filters in this lower-dimensional space. The decoder model is designed to up-sample these modified lower-dimensional features to higher-dimensional data equal to the dimensions of original input image x. The decoder model takes the modified feature vector as an input and produces an altered output image $x'$ which is the counterfactual of the original input image x. }
\begin{equation}
x' = D\bigg(C\big(x, F_{MC_i}, F_{MI_i}\big)\bigg).
\end{equation}
The decoder generates the image which reflects the filter-level changes made in the latent space vector as shown in Figure 1. The reconstructed image by the decoder provides visual explanations of the features-modification made by the counterfactual and contrastive filters.This reconstructed image represents a plausible visual explanation that aligns with the internal decision-making process of the model. \textcolor{black}{The procedure describing the overall approach is presented in Algorithm \ref{algo:1}.}

The proposed approach allows us to gain insights into the influence of specific filters on the model's decision-making process and the generated visual explanations provide a better understanding of how the model arrives at its decisions.

\begin{algorithm}[htbp]
	\caption{\textcolor{black}{Steps to generate counterfactual visual explanations.}} 
	\label{algo:1}
	\textbf{Input: }{Image $I$, Classifier model $C$, Counterfactual explanation model $CFE$, target class $\hat{c}$, train dataset $T$}
	\begin{algorithmic}[0]
    
    \State Step 1. Train a decoder
     \Procedure{TrainDecoder}{$C$}
      \State Train the decoder on train dataset $T$ using features from classifier $C$
         \Indent
    	       \State{$\min_{D}\frac{1}{n}\sum_{i=1}^{n}|x_{i}-D(C_{conv}(x_{i}))|_{1}$}
        \EndIndent
    \EndProcedure

    \State {Step 2. Generate contrastive and counterfactual explanation using CFE model for input image $I$}
     \Procedure{GenerateExplanation}{$I$, ${CFE}$}
        \State $F_{MI_i}, F_{MC_i} = \text{CFE}(I,C,\hat{c})$
     \EndProcedure

    \State {Step 3. Alter filters in classifier}
     \Procedure{AlterFilters}{$C$, $F_{MI_i}$, $F_{MC_i}$}
        \State {Generate feature vector $g$ and predicted class $c$ using classifier $C$}: \State $g,c = C(I)$
        \State $\hat{c} = h( g \circ F_{MC_i})$ \Comment{alter prediction with just MC filters enabled}
        \State $\hat{c} = h( g + F_{MI})$ \Comment{alter prediction with just updated MI filters}
        \State where $h$ represents the classification (fully-connected and softmax) layers of $C$
     \EndProcedure     

    \State {Step 4. Use trained decoder to generate visual explanation by reconstructing input $I$ with modified classifier}
     \Procedure{VisualExplanation}{$D$, $C$, $F_{MI_i}$, $F_{MC_i}$}
        \State $I' = D\bigg(C\big(I, F_{MC_i}, F_{MI_i}\big)\bigg)$
     \EndProcedure         

	\end{algorithmic}
	\textbf{Output: }{$I'$} \Comment{Reconstructed input image with counterfactual and contrastive features using the modified classifier}
\end{algorithm}

\section{Results}
\label{sec:four}

This section presents the results and discussion of the proposed FCVE method. For the evaluation of the proposed FCVE method, we used MNIST \cite{lecun1998mnist} and Fashion-MNIST (FMNIST) \cite{xiao2017fashion} datasets and compared with related counterfactual explanation methods including ExpGAN \cite{samangouei2018explaingan}, CEM \cite{dhurandhar2018explanations}, CVE \cite{pmlr-v97-goyal19a}, and C3LT \cite{khorram2022cycle}. In section \ref{sec:four-1} and \ref{sec:four-2}, we present a visual comparison of these methods on the two datasets, followed by quantitative analysis presented in Section \ref{sec:four-3}. 

\textcolor{black}{To evaluate faithfulness of the explanations provided by the proposed method, we refer the reader to \cite{TARIQ2022109901} that used the class recall metric to demonstrate that the identified counterfactual and contrastive filters are faithful to their respective classes. In  \cite{TARIQ2022109901}, it was shown that disabling around 31–44 most imported filters of a class (out of 512 total filters) resulted in a significant decrease in the class recall, whereas the overall model accuracy was reduced by just 2\%–3\%. On the contrary, it was shown that randomly disabling the same number of filters had a negligible effect on class recall. This shows that the counterfactual and contrastive filters predicted by the CFE model represent features exclusive to a particular class and disabling them slightly affects the overall model accuracy while significantly reducing the particular class’s recall score, thus demonstrating the faithfulness of the detected filters used in the decision-making process of the classifier. In the proposed work, we mainly focus on the visual aspect of faithful explainable approach. }

\begin{figure*}[h!]
\begin{center}
    \includegraphics[width=1.0\linewidth]{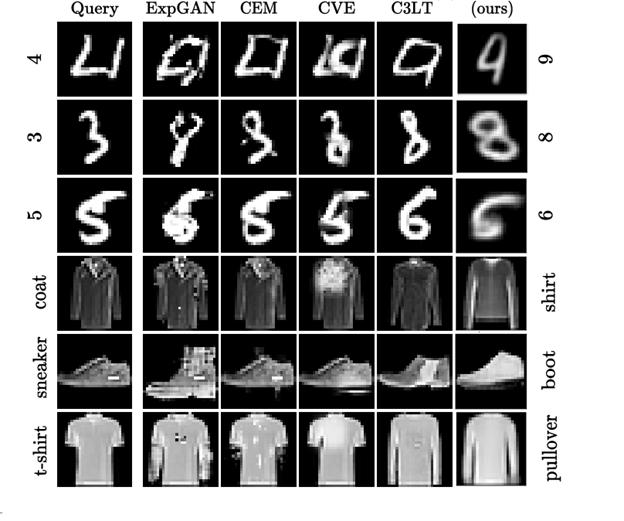}
\end{center}
   \caption{Visual comparison of counterfactual explanation methods. \textcolor{black}{The first column shows the query images from MNIST and FMNIST, while the other five columns display the counterfactuals generated by ExpGAN \cite{samangouei2018explaingan}, CEM \cite{dhurandhar2018explanations}, CVE \cite{pmlr-v97-goyal19a}, C3LT \cite{khorram2022cycle}, and our proposed model (FCVE), respectively. The proposed method generates counterfactuals by manipulating the internal activations of the model, resulting in counterfactuals that are more meaningful and realistic compared to other methods.}}
\label{fig:2}
\end{figure*}

\subsection{Visual results comparison with related methods}
\label{sec:four-1}

Figure \ref{fig:2} represents a comparison of the counterfactual explanation results. The first column shows the query images from MNIST and FMNIST, while the other five columns display the counterfactuals generated by ExpGAN \cite{samangouei2018explaingan}, CEM \cite{dhurandhar2018explanations}, CVE \cite{pmlr-v97-goyal19a}, C3LT \cite{khorram2022cycle}, and our proposed model (FCVE), respectively. Our method generates counterfactuals by manipulating the internal activations of the model, resulting in counterfactuals that are more meaningful and realistic compared to other methods. We ensured that the source and target classes were selected to maintain the counterfactual proximity property.

Among the baseline models, the results of C3LT are somewhat interpretable but mainly unrealistic. The counterfactuals obtained from C3LT are adversarial to the target classes (e.g., generating 8 and 9 from 3 and 4, respectively, and a shirt from a coat). The counterfactuals obtained from ExpGAN are not smooth (e.g., 9, 6, and pullover). The counterfactuals from CEM and CVE are unrecognizable (e.g., 9 and 8) or mostly unchanged (e.g., 6, short, boot, and pullover). In contrast, the counterfactuals generated by our method are easily recognizable and more realistic.

\subsection{Qualitative analysis of proposed method}
\label{sec:four-2}
This section presents an additional qualitative analysis of the proposed FCVE methods in terms of generating plausible visual counterfactuals for MNIST and FMNIST datasets.

\subsubsection{MNIST counterfactuals}

\begin{figure}[hbt]
\begin{center}
    \includegraphics[width=1.0\linewidth]{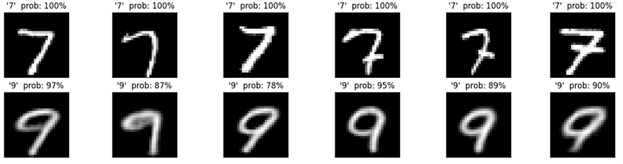}
\end{center}
   \caption{Plausible counterfactuals generated for digit seven as a source class and digit nine as target class. \textcolor{black}{The proposed method finds the minimal changes to neuron activations such that the input of one class is transformed into another.}}
\label{fig:3}
\end{figure}

Figure \ref{fig:3} displays the counterfactuals generated for the digit seven as the source class and the digit nine as the target class. Despite the non-identical writing styles of the input images for the same digit, our method successfully generates plausible counterfactuals. This ability to generate counterfactuals indicates that the model has learned the underlying data patterns and can generalize well.

The first three input images (1st row) of the digit seven vary in writing style compared to the last three images, which include an extra intersection line. While humans can easily differentiate between these variations of the digit seven, it can be a challenging task for an algorithm to identify such subtle changes.

\begin{figure}[hbt!]
\begin{center}
    \includegraphics[width=1.0\linewidth]{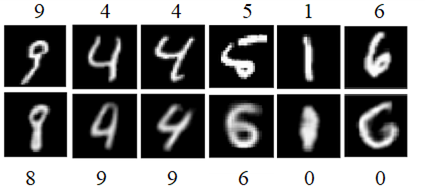}
\end{center}
   \caption{Counterfactuals generated for random source and target classes of MNIST dataset. \textcolor{black}{Similar to Fig. \ref{fig:3}, the proposed method finds the minimal changes to neuron activations such that the input of one class is transformed into another.}}
\label{fig:4}
\end{figure}

Figure \ref{fig:4} displays the counterfactuals generated for randomly selected source and target classes of the MNIST dataset. The input images in the first row (i.e., 9, 4, 4, 5, 1, and 6) are chosen randomly, while the images in the second row (i.e., 8, 9, 9, 6, 0, and 0) represent the counterfactuals generated by our model. Our model aims to generate counterfactuals by adding or subtracting features from the original input image. For example, the first counterfactual (8) is obtained by adding a line to the input image (9). Similarly, the counterfactuals of 0, 9, and 6 are generated by the same principle from the input images 1, 4, and 5, respectively. Additionally, a counterfactual of 0 is obtained by removing a portion of 6.

\subsubsection{FMNIST counterfactuals}

\begin{figure}[hbt!]
\begin{center}
    \includegraphics[width=1.0\linewidth]{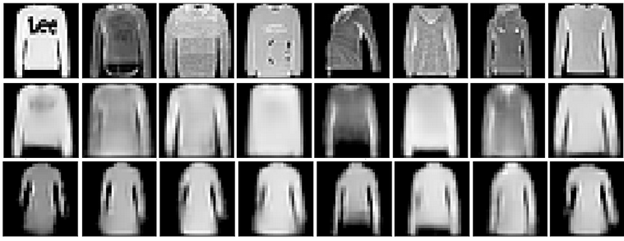}
\end{center}
   \caption{Plausible counterfactuals generated for the FMNIST dataset. \textcolor{black}{The first row is the source class “Pullover”. Second and third rows are target classes of “Dress” and “Coat” into which the source image is transformed into by altering the filter activations.}}
\label{fig:5}
\end{figure}

Figure \ref{fig:5} displays the results of the counterfactual visual explanations obtained using the proposed method. In this analysis, the source class is ``Pullover," represented by the images in the first row. The goal is to transform these pullover images into counterfactual representations of the target classes, which are ``Dress" and ``Coat" displayed in the second and third rows, respectively. These counterfactuals are generated by the FCVE model, utilizing the source class image as a starting point.

It can be seen that the proposed method successfully modifies the source class images to generate plausible visual counterfactuals that accurately represent the target classes. The generated counterfactuals exhibit visual characteristics and features associated with the respective target classes, showcasing the effectiveness of the FCVE model in capturing and manipulating the underlying data patterns.
 
\begin{figure}[hbt!]
\begin{center}
    \includegraphics[width=1.0\linewidth]{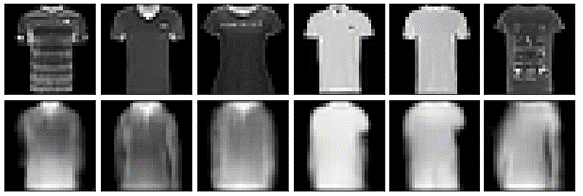}
\end{center}
   \caption{Counterfactual generation from visually identical classes of ``T-shirt" (source class, row 1) and ``Pullover" (target class, row 2) in FMNIST. \textcolor{black}{The proposed method effectively transforms the t-shirts into pullovers with distinctive features, such as long sleeves.}}
\label{fig:6}
\end{figure}

\begin{figure}[hbt!]
\begin{center}
    \includegraphics[width=1.0\linewidth]{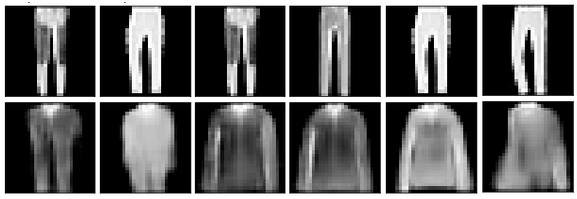}
\end{center}
   \caption{Counterfactual generation from visually non-identical classes of ``Trouser" (source class, row 1) and ``Shirt" (target class, row 2) in FMNIST. \textcolor{black}{Despite the differences between the source and target classes, the proposed method produces realistic target class images by transforming the source images.}}
\label{fig:7}
\end{figure}

Figures \ref{fig:6} and \ref{fig:7} showcase additional examples of counterfactual image generation from visually identical and non-identical classes, respectively.
In Figure \ref{fig:6}, the first row comprises actual images of t-shirts from the FMNIST dataset, while the second row displays the counterfactuals generated by our proposed model, for the target class of ``Pullover". The source class (t-shirts) and target class (pullover) belong to visually similar categories, and the proposed method effectively transforms the t-shirts into pullovers with distinctive features, such as long sleeves. It is worth noting that the model accurately captures the shape of the target class while sacrificing some finer details, such as patterns on the t-shirts. This suggests that the shape is a more crucial feature than the specific patterns when differentiating between these classes.

Similarly, in figure \ref{fig:7}, the counterfactuals generated by our proposed model are presented, focusing on visually diverse source and target classes. In this case, the source class is ``Trouser", while the target class is ``Shirt". These classes exhibit noticeable visual differences in terms of shape, texture, and overall appearance. Despite the visual disparity between the source and target classes, our proposed model consistently produces realistic target class images by transforming the source images. This demonstrates the effectiveness of our approach in generating accurate and visually coherent counterfactual representations.

\subsection{Quantitative comparison}
\label{sec:four-3}

This section provides a quantitative analysis of the proposed FCVE method and compares it with existing methods in terms of the proximity measure and Fréchet Inception Distance (FID).
\textcolor{black}{\subsubsection{Proximity}
Proximity property explanations the counterfactuals, meaning Faithful to the original instance. The generated counterfactual explanations are considered the best as they are closest to the original instance.  Proximity is the mathematical formula to quantify the closeness of two instances (query image and counterfactual) using L1 distance. Satisfying, this minimal feature change property to generate counterfactual examples, the proximity metric can be defined mathematically in terms of distance function as,
\begin{equation}
    \text{proximity} = \frac{1}{N} \sum_{i=1}^{N} \frac{\text{dsict}(x_i - x_i')}{CHW}
\end{equation}
where $x_i$ and $x'_i$ represent the ith query image and counterfactual example from the set being evaluated, and $C$, $W$ and $H$ are the channels count, width, and height of the query image, respectively.
The Lower levels of proximity suggest methods that produce counterfactuals that are closer to the original data points.
\subsubsection{Plausibility}
Plausibility property depicts the counterfactual explanations are realistic, feature values are coherent to the domain set. The feature values of counterfactuals should not be an outlier in consideration with domain set. Enhancing trust in the explanation is facilitated by plausibility. The approach we are using to check plausibility is FID score calculation. It is feature-wise subtraction of the query images and their respective counterfactuals. Plausibility contributes to the robustness and stability of counterfactual explanations.
The formula for calculating the FID is as follows:
\begin{equation}
FID = \lvert \mu - \mu' \rvert^2 + \text{Tr} \left( x + x' - 2\sqrt{x . x') } \right)
\end{equation}
Where $\mu,\mu' , x, x', \text{Tr}(\cdot), \lvert \cdot \rvert^2 $ denotes mean feature vectors of the real and generated image distributions, the covariance matrices of the real and generated image distributions, the trace of a matrix and the squared Euclidean norm.
the FID metric measures the similarity between the generated images and real images, focusing on the distribution of features. A lower FID score indicates better-quality images and greater realism.}

Table \ref{tab:comparison} presents a comparison of the counterfactual explanation methods based on both the proximity and FID metrics. These metrics were obtained from various baseline models, including ExpGAN \cite{samangouei2018explaingan}, CEM \cite{dhurandhar2018explanations}, CVE \cite{pmlr-v97-goyal19a}, and C3LT \cite{khorram2022cycle}. From the table, it is evident that the proposed FCVE method achieves a significantly lower FID score compared to the compared methods. This result indicates that the proposed method generates high-quality counterfactuals that closely resemble the real data, demonstrating its effectiveness in generating realistic and meaningful counterfactual explanations.

\begin{table*}[htbp]
\scriptsize
\centering
\caption{Comparison of counterfactual explanation methods on MNIST and FMNIST datasets based on proximity and FID scores.}
\label{tab:comparison}
\begin{tabular}{lcccccccccc}
\toprule
Method & \multicolumn{2}{c}{ExpGAN \cite{samangouei2018explaingan}} & \multicolumn{2}{c}{CEM \cite{dhurandhar2018explanations}} & \multicolumn{2}{c}{CVE \cite{pmlr-v97-goyal19a}} & \multicolumn{2}{c}{C3LT \cite{khorram2022cycle}} & \multicolumn{2}{c}{FCVE (our)} \\
\cmidrule(lr){2-3} \cmidrule(lr){4-5} \cmidrule(lr){6-7} \cmidrule(lr){8-9} \cmidrule(lr){10-11}
& MNIST & FMNIST & MNIST & FMNIST & MNIST & FMNIST & MNIST & FMNIST & MNIST & FMNIST \\
\midrule
Proximity & 0.074 & 0.135 & 0.016 & 0.013 & 0.055 & 0.054 & 0.072 & 0.116 & 0.098 & 0.198 \\
FID	& 41.12	& 76.52	& 50.03	& 96.87	& 47.53	& 83.77	& 22.83	& 62.31	& 0.50	& 2.02\\
\bottomrule
\end{tabular}
\end{table*}

\section{Conclusion}
\label{sec:five}

The development of explainable AI techniques plays a crucial role in addressing the transparency and interpretability challenges associated with deep learning models in computer vision. While significant progress has been made, existing methods still face limitations in providing convincing explanations that are easily understandable to non-experts and accurately capture the intrinsic decision-making processes of the models. 

To overcome these challenges, we have proposed a counterfactual explanation (CE) model that aims to strike a balance between plausibility and faithfulness. Our model generates visual explanations that are not only easy to comprehend but also faithfully represent the model's internal decision-making process. Importantly, these explanations are generated by making minimal changes to the original images, without altering the pixel data.

Instead of relying solely on pixel-level manipulations, our approach identifies and leverages the internal concepts and filters learned by the model. By understanding and manipulating these internal representations, our model produces plausible counterfactual explanations that reflect the model's underlying decision-making process, making the provided explanation faithful to the model.

Through qualitative and quantitative analysis, we have demonstrated the effectiveness of our proposed FCVE method. The qualitative analysis highlights the close resemblance between the generated counterfactuals and the original data instances, indicating the high quality of the explanations. Furthermore, the quantitative analysis using Fréchet Inception Distance (FID) scores confirms that our method outperforms the baseline models in generating realistic and diverse counterfactuals.

Future research directions could focus on extending the proposed method to other domains and exploring additional evaluation metrics to further validate the effectiveness of counterfactual explanations in different contexts.

\bibliographystyle{plainnat}
\bibliography{bibliography}






\end{document}